%% file: main.tex
\documentclass[10pt,twocolumn,letterpaper]{article}

\usepackage[pagenumbers]{wacv} 
\usepackage{times}
\usepackage{epsfig}
\usepackage{graphicx}
\usepackage{amsmath}
\usepackage{amssymb}

\usepackage{multirow}
\usepackage[table,xcdraw]{xcolor}
\usepackage{pifont}
\usepackage{tabularx} 
\usepackage{float}


\definecolor{wacvblue}{rgb}{0.21,0.49,0.74}
\usepackage[pagebackref,breaklinks,colorlinks,allcolors=wacvblue]{hyperref}

\def\wacvPaperID{1515} 
\def\confName{WACV}
\def\confYear{2026}

\title{PromptGAR: Flexible Promptive Group Activity Recognition}

\author{
Zhangyu Jin$^{1}$ \hspace{12pt} Andrew Feng$^{1}$ \hspace{12pt} Ankur Chemburkar$^{1}$ \hspace{12pt} Celso M. De Melo$^{2}$\\
\\
$^1$ University of Southern California, Institute for Creative Technologies \\
$^2$ Army Research Laboratory \\
{\tt\small \{zjin,feng,achemburkar\}@ict.usc.edu, celso.m.demelo.civ@army.mil}
}

\begin{document}
\maketitle


\begin{abstract}
We present \textbf{PromptGAR}, a novel framework for Group Activity Recognition (GAR) that offering both input flexibility and high recognition accuracy. 
The existing approaches suffer from limited real-world applicability due to their reliance on full prompt annotations, fixed number of frames and instances, and the lack of actor consistency.
To bridge the gap, we proposed PromptGAR, which is the first GAR model to provide input flexibility across prompts, frames, and instances without the need for retraining.
We leverage diverse visual prompts—like bounding boxes, skeletal keypoints, and instance identities—by unifying them as point prompts. A recognition decoder then cross-updates class and prompt tokens for enhanced performance.
To ensure actor consistency for extended activity durations, we also introduce a relative instance attention mechanism that directly encodes instance identities.
Comprehensive evaluations demonstrate that PromptGAR achieves competitive performances both on full prompts and partial prompt inputs, establishing its effectiveness on input flexibility and generalization ability for real-world applications.
\end{abstract}

\section{Introduction}

Group Activity Recognition (GAR) \cite{choi2009they} is fundamentally important for video and event understanding, and it is widely used in areas such as video analytics, human computer interaction, and security systems. GAR processes videos and annotations including bounding boxes, skeletons, ball trajectories, and optical flows to determine the group activity label. Building on prior research \cite{zhou2022composer, li2025skeleton, gavrilyuk2020actor, li2021groupformer, han2022dual}, we aim to not only enhance group activity recognition accuracy but also to create a more flexible architecture handling diverse prompt, frame, and instance inputs.
 
Although numerous models have been proposed, high-performance and flexible group activity recognition continues to be challenging due to the following difficulties:

\begin{figure*}[h]
\centering
\includegraphics[width=0.8\linewidth]{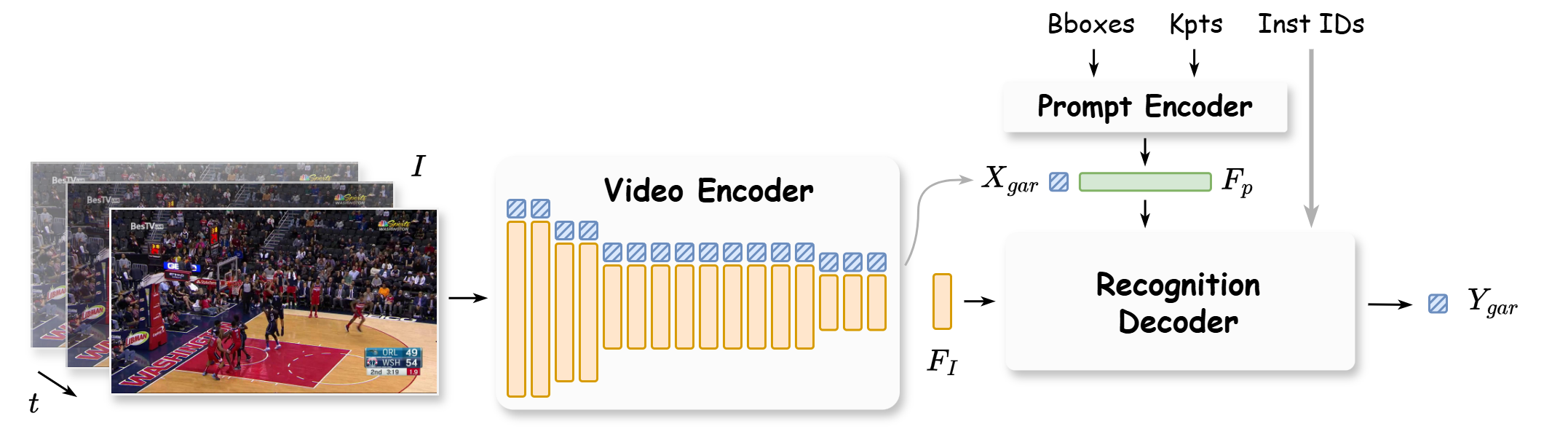}
\vspace{1.0em}
\caption{\textbf{PromptGAR Architecture}. A sequence of frames $I$ is processed by the video encoder, yielding RGB features $\textbf{F}_I$ and GAR class token $\textbf{X}_{gar}$. Prompts, such as bounding boxes and skeletal keypoints, are transformed to prompt tokens $\textbf{F}_p$ by the prompt encoder. These tokens, along with instance identities, are then fed into the recognition decoder to get group activity prediction $\textbf{Y}_{gar}$.}
\label{fig:promptgar_architecture}
\end{figure*}

\textbf{Requirement for Full Prompts}. 
Current group activity recognizers \cite{gavrilyuk2020actor, li2021groupformer, han2022dual} rely on full annotations to achieve strong performance at test time.
It is widely acknowledged that obtaining accurate annotations is difficult. Even though using state-of-the-art object detectors \cite{zong2023detrs, su2023towards}, trackers \cite{stanojevic2024boosttrack++, yi2024ucmctrack}, and pose estimators \cite{xu2022vitpose, geng2023human}, the lower quality annotations still exist, such as missed detections, redundant boxes with low confidence scores, player ID switching, player ID reassignment upon reappearance, and so on.
In real-world scenarios, manually correcting these annotations at test time is often impractical.
Instead of being forced to use potentially inaccurate prompts during test time, users in real-world scenarios want the flexibility to choose from full, partial, or no prompts.
However, current GAR architectures \cite{zhou2022composer, li2025skeleton, gavrilyuk2020actor, li2021groupformer, han2022dual, yuan2021spatio, pei2023key, pramono2020empowering} either do not support partial prompts, or they demand retraining to perform reasonably.
So we design our model for the input flexibility: it is trained with full prompts like prior methods, but it delivers competitive results with full prompts and still performs quite well with fewer or no prompts at test time, without retraining.

\begin{figure}[h]
\centering

\makebox[\columnwidth][c]{
    \includegraphics[width=1.0  \columnwidth]{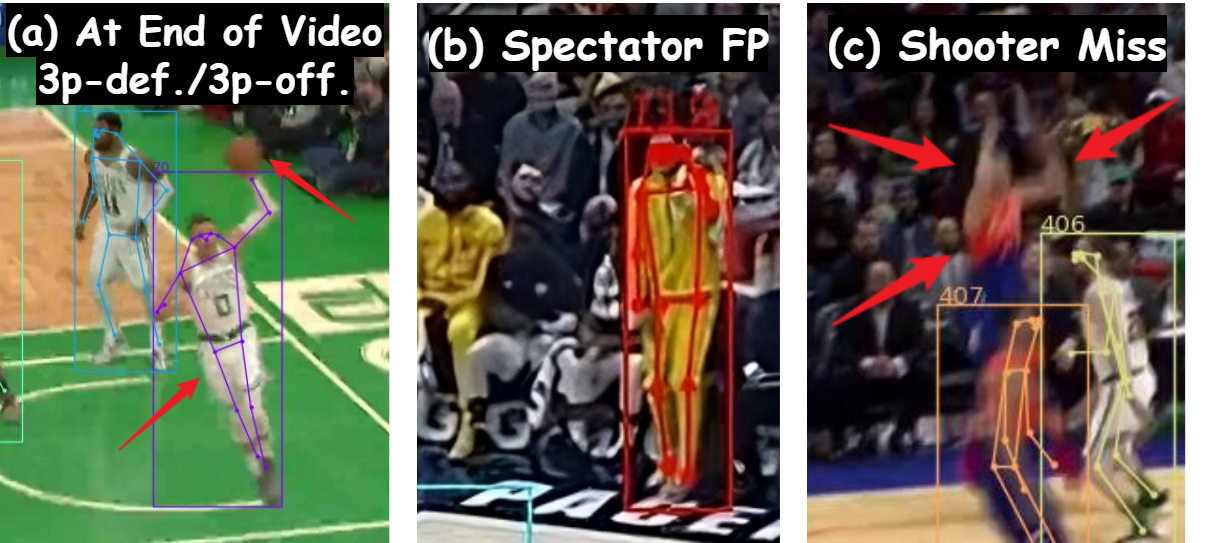}
}
\caption{Necessity of flexible frames: (\textit{a}) offense or defense relies on which team gets the ball at the end, and similar key moments can be anywhere in the video. Necessity of flexible instances: (\textit{b}) false-positives and (\textit{c}) false-negatives make player counts unfixed.}
\label{fig:qualitative_analysis}

\end{figure}

\textbf{Requirement for Fixed Frames and Instances}. 
Most current GAR methods \cite{gavrilyuk2020actor, li2021groupformer, han2022dual} are limited to a fixed number of frames and instances as input. 
Those rigid requirements lead to significant performance degradation in real-world scenarios:
(1) As seen in Fig. \ref{fig:qualitative_analysis}-a, offense or defense relies on who gets the ball at the end. 
Such key moment can be anywhere in the video, so previous fixed-frame models may miss it depending on how the frames are sampled from a long sequence. 
(2) Similarly, in Fig. \ref{fig:qualitative_analysis}-b and c, real-world data often contains missing actors in annotations or false positive detections of spectators as players.
Prior fixed-instances models requires a fixed input shape, causing runtime errors or requiring arbitrary padding/truncation that degrades performance.
In contrast, our model is designed to accept a flexible number of frames and instances, and it achieves reliable performance without requiring retraining.

\textbf{Lacking Actor Consistency}. 
Recent GAR approaches \cite{zhang2024bi, li2025skeleton, zhou2022composer} rely on a fixed player order. Their performance degrades when this order changes.
Even though player order in the testing set of either Volleyball \cite{ibrahim2016hierarchical} or NBA \cite{yan2020social} is not supposed to be known, that order is still fixed for both validation and testing in these methods.
Consequently, among all epochs, they pick the checkpoint with the best performance under this specific player order. 
However, in real-world scenarios, such player order is entirely unknown.
A robust GAR model, therefore, should give identical performance regardless of the input player order.
In our work, we address this by introducing a relative instance attention mechanism that only encodes whether two players are the same or not. 
This design ensures that player order does not affect our model's performance at all.

Considering the above challenges and motivations, we present \textbf{PromptGAR}, a transformer-based group activity recognition framework that leverages diverse visual prompts (i.e., bounding boxes, skeletal keypoints, instance identities) to achieve high group activity recognition accuracy and input flexibility. 
(a) \textbf{Flexible Prompts}. It adapts to varying prompt availability, namely full prompts, partial prompts, and no prompts. When comprehensive annotations are present, the full prompt inputs would maximize performance. If only simpler annotations like bounding boxes and instance identities are available, our model still gives reliable results. Even with no annotations, it can also provide reasonable results using only the raw video. This adaptability makes it suitable for diverse real-world scenarios where annotation quality varies.
(b) \textbf{Flexible Frames}. The method has temporal flexibility in accepting videos of varying lengths and frame rates. Furthermore, it effectively recognizes group activities regardless of their temporal location within the video, accommodating both instantaneous actions and sequential events. 
(c) \textbf{Flexible Instances}. It is designed to handle varying numbers of actors within a scene. 
Therefore in an ideal scenario, our model can leverage detailed annotations of all individuals for enhanced accuracy. On the other hand, it can also maintain robust performance even when there are missing annotations for certain actors.
(d) \textbf{Without Retraining}. After training with full prompts, PromptGAR automatically gains the flexibility for input prompts, frames, and instances during inference. This significantly enhances its practicality and broadens its applicability to diverse scenarios.

Our implementation is inspired by the Segment Anything \cite{kirillov2023segment}. We unify bounding boxes and skeletal keypoints as point prompts and employing a two-way decoder for cross-updating class and prompt tokens.
\textbf{Firstly}, the input flexibility is implemented through several design features: MViTv2's relative positional embedding for variable length of RGB frames, 
depth-wise prompt pooling to accommodate flexible temporal and prompt dimensions, 
and a robust head that maintains classification stability even when no prompts are available during inference.
\textbf{Secondly}, to effectively acquire actor consistency, we introduce relative instance attention, which directly encodes instance IDs. The encoding ensures that the output will remain invariant to instance ID transitions.

Based on the above technical contributions, we evaluate our model under Volleyball \cite{ibrahim2016hierarchical} and NBA \cite{yan2020social} datasets, PromptGAR produces competitive results compared to state-of-the-art GAR methods, showcasing its strong recognition capabilities. Additionally, we demonstrated its flexibility to handle varying prompt inputs, frame sampling, and instance counts while maintaining robust performance without retraining. 
Our main contributions are as follows.

(\textit{a}) An effective group activity recognition architecture that achieves input flexibility across prompts, frames, and instances without retraining.

(\textit{b}) A relative instance attention module for encoding instance identities and ensuring actor consistency.

\section{Related Work}

\subsection{Group Activity Recognition}
GAR \cite{choi2009they} has attracted attention due to its massive success in a variety of real-world applications. Earlier techniques relied heavily on handcrafted features \cite{choi2009they, lan2011discriminative, hajimirsadeghi2015visual, cheng2010group, choi2011learning, choi2012unified, nabi2013temporal} or AND-OR graphs \cite{amer2013monte, amer2012cost, shu2015joint}. Recently, methods based on neural network architectures have been widely studied because of their ability to effectively extract features and fuse various visual prompts.

\textbf{Various Prompts for GAR.} 
(a) \textit{RGB frames and bounding boxes}. Existing work either directly crops people from scenes \cite{yan2020higcin} or applies ROIAlign \cite{he2017mask} to represent actors from extracted feature maps \cite{yuan2021spatio, pramono2020empowering, li2021groupformer}. 
(b) \textit{Skeletons}. Several works \cite{thilakarathne2022pose, zhou2022composer, duan2023skeletr, li2025skeleton} explore using human skeleton joints as inputs to avoid substantial computational resources and discrepancies in  background and camera settings in video-based approaches. 
(c) \textit{Optical flows}. Neighboring RGB frames can produce optical flow images \cite{zach2007duality}. These flows are then concatenated with RGB frames to create dense inputs to introduce pixel-wise motion information across the temporal dimension \cite{pramono2020empowering, li2021groupformer, han2022dual}. A similar idea is also applied in Composer \cite{zhou2022composer} for skeletal information by computing the temporal difference of coordinates in two consecutive frames. MP-GCN \cite{li2025skeleton} calculated joint motions and bone motions as extra sparse inputs.
(d) \textit{Balls}. Group labels in the NBA dataset \cite{yan2020social}, such as `\textit{2p-succ}' and `\textit{3p-succ}', are closely related to ball positions; therefore, Composer \cite{zhou2022composer} and MP-GCN \cite{li2025skeleton} treat balls as additional inputs. However, it is difficult to leverage these specific model designs for general scenarios such as non-sports activity recognition.
(e) \textit{Tracked instance identities}. Recent approaches \cite{ yuan2021spatio, li2021groupformer} use optical flows to describe only short-term motions without explicitly encoding instance identities to preserve actor consistency. We introduce a relative instance identities encoding mechanism to handle this long-neglected prompt.

\begin{figure}[h]
\centering

\makebox[\columnwidth][c]{
    \includegraphics[width=1.1\columnwidth]{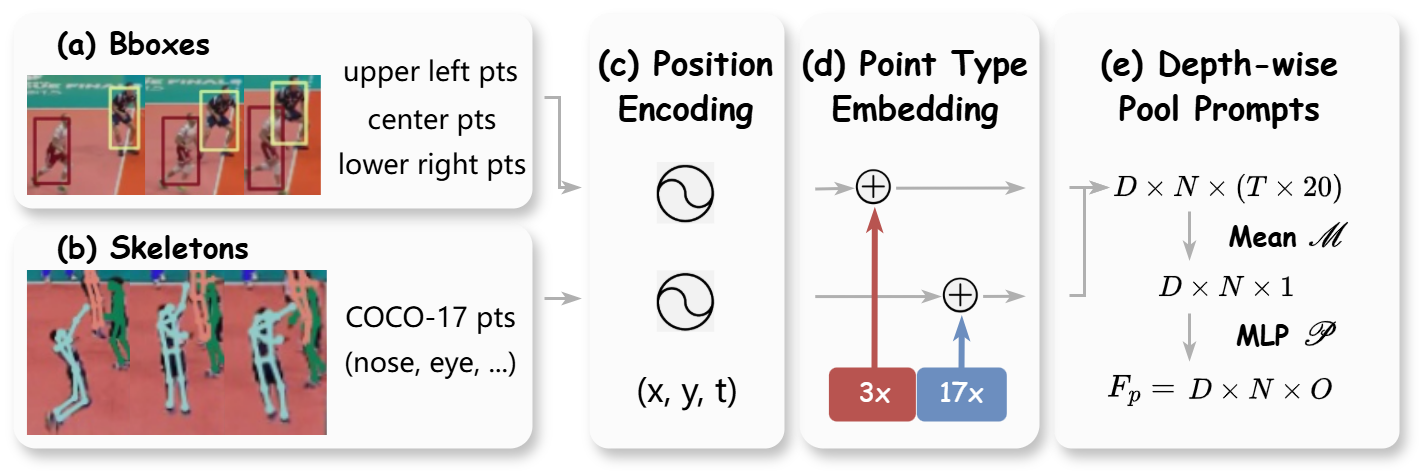}
}
\caption{\footnotesize \textbf{Prompt Encoder}. (\textit{a}) Bounding boxes are represented by 3 points (upper-left, center, lower-right). (\textit{b}) Skeletal keypoints consist of 17 points. (\textit{c}) Positional encoding captures both spatial and temporal coordiates. (\textit{d}) Point types are distinguished using learnable embeddings. (\textit{e}) Depth-wise prompts pooling reduces temporal and type dimensions to 1, then up-projects to the number of pooled prompts $O$.}
\label{fig:prompt_encoder}

\bigskip

\makebox[\columnwidth][c]{
    \includegraphics[width=1.1\columnwidth]{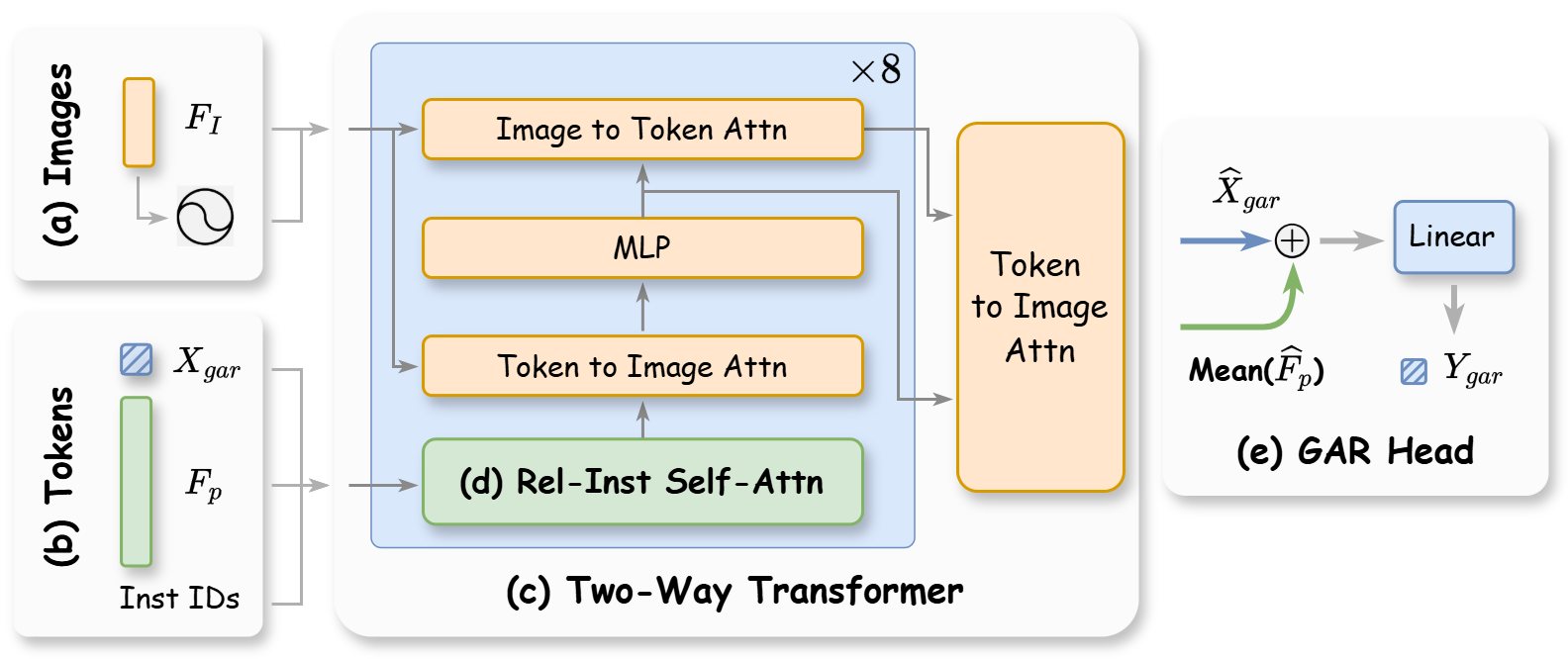}
}
\caption{\footnotesize \textbf{Recognition Decoder}. The decoder processes (\textit{a}) RGB features $\textbf{F}_I$ with positional embeddings and (\textit{b}) the GAR class token $\textbf{X}_{gar}$ and prompt tokens $\textbf{F}_p$. (\textit{c}) The Two-Way Transformer performs cross-updating between these features. (\textit{d}) Relative instance self-attention, using instance identities, ensures actor consistency. (\textit{e}) The GAR head takes updated GAR class token $\widehat{\textbf{X}}_{gar}$ and average of updated prompt tokens $\widehat{\textbf{F}}_p$ to predict the group activity label $\textbf{Y}_{gar}$.}
\label{fig:recognition_decoder}
\end{figure}

\textbf{Input Flexibility for GAR}. Due to the model design choices, recent works \cite{zhou2022composer, li2025skeleton, gavrilyuk2020actor, li2021groupformer, han2022dual} have less input flexibility in group activity recognition. 
(a) \textit{Flexible prompts}. A common limitation among GAR methods \cite{gavrilyuk2020actor, li2021groupformer, han2022dual} that process RGB frames and bounding boxes is their dependence on ROIAlign \cite{he2017mask} for actor feature extraction, making them incompatible with scenarios lacking bounding box prompts. Likewise, GroupFormer's fixed input channel setting \cite{li2021groupformer}, achieved through concatenating RGB and skeleton features, prevents its use when skeleton data is unavailable. 
(b) \textit{Flexible frames}. Dual-AI \cite{han2022dual} demonstrates some flexibility by applying different frame sampling strategies during training and testing, such as 3 frames for training and 9 or 20 frames for testing on the Volleyball \cite{ibrahim2016hierarchical} and NBA \cite{yan2020social} datasets, respectively. However, these frame counts remain significantly lower than the total available frames (41 for Volleyball, 72 for NBA) in each clip. This restricted frame sampling flexibility limits its potential for achieving higher performance.
(c) \textit{Flexible instances}. Composer \cite{zhou2022composer} relies on normalizing joint coordinates with statistics derived from the complete clip. However, missing instances significantly alter these statistics, shifting the mean and standard deviation far from their expected distribution. 
Notably, our PromptGAR maintains high performance across inputs with flexible prompts, frames, and instances without the needs for retraining.

\section{Method}
As shown in Fig. \ref{fig:promptgar_architecture}, PromptGAR is an end-to-end prompt-based framework for group activity recognition. It takes a sequence of frames and corresponding prompts as inputs and outputs the group activity label. In the following subsection, we first introduce how to encode videos and visual prompts in $\S$\ref{sec:prompt_encoding}. Then we illustrate how to fuse various visual prompts and spatial-temporal information in $\S$\ref{sec:recognition_decoding}.

\subsection{Visual Inputs Encoding \label{sec:prompt_encoding}}

PromptGAR supports three kinds of inputs: input frames $\textbf{I}$, bounding box prompts $\textbf{F}_{box}$, and skeleton prompts $\textbf{F}_{kpt}$.

\textbf{Video Encoding}. 
While existing works \cite{yuan2021spatio, han2022dual} have shown promises in their power to model RGB features, they focus on images rather than videos. Other studies \cite{pramono2020empowering, li2021groupformer} chose I3D \cite{carreira2017quo}, but they require optical flows as extra dense guidance. To address those issues, the sequence of frames $\textbf{I}$ is processed through MViTv2 \cite{li2022mvitv2}, to extract multi-scale feature tokens $\textbf{F}_I$ and the GAR class token $\textbf{X}_{gar}$.

\textbf{Point Prompts}. 
As shown in Fig. \ref{fig:prompt_encoder}, all of the aforementioned prompts are formulated as point prompts: 
(a) Bounding boxes are formatted in three points, including upper-left, center, and lower-right points, similarly as in \cite{kirillov2023segment}. 
(b) Human skeletons follow the definition of COCO keypoints \cite{lin2014microsoft}. 
In general, one point is uniquely described by five attributes:
$$(x,y,t,p,\textit{ID})$$
where $x,y,t$ are the positions along the width, height, and temporal axes, $p$ is the point type, and $\textit{ID}$ denotes the instance identity that the point belongs to.
We encode these attributes in the following ways.
\textit{Firstly}, while the original Fourier embedding \cite{tancik2020fourier} only maps the spatial coordinate $(x,y)$ to the corresponding feature dimensions, we extend its capability to incorporate temporal location $t$.
$$\gamma (\textbf{v}) =  \Big[\cos\Big(2\pi \textbf{B}(2 \textbf{v}-1)\Big), \sin\Big(2\pi \textbf{B}(2 \textbf{v}-1)\Big) \Big]^T$$
where $\textbf{v}:=[x,y,t]^T\in[0,1]^3$ is the spatial-temporal coordinate, and $\textbf{B}\in\mathbb{R}^{\lfloor D/2\rfloor\times3}$ is sampled from a Gaussian distribution $\mathcal{N}(0,1)$.
The same positional encoding is also applied to RGB features $\textbf{F}_I$.
\textit{Secondly}, we employ learned embeddings for each prompt type $p$, similar to \cite{kirillov2023segment}. 
\textit{Thirdly}, $\textit{ID}$ is encoded by relative embeddings, with details in $\S$ \ref{sec:recognition_decoding}.

\textbf{Depth-wise Prompts Pooling}. 
As described in Fig. \ref{fig:prompt_encoder}-(e), after obtaining prompt features $\textbf{F}_{box}$ and $\textbf{F}_{kpt}$, we employ a mean pooling operator $\mathcal{M}$ and a MLP projector $\mathcal{P}$. They perform along temporal and type dimensions.
$$\textbf{F}_p=(\mathcal{P}\circ\mathcal{M})([\textbf{F}_{box};\textbf{F}_{kpt}])$$
where $\textbf{F}_p\in\mathbb{R}^{D\times N\times O}$, $\textbf{F}_{box}\in\mathbb{R}^{D\times N\times (T\times 3)}$ and $\textbf{F}_{kpt}\in\mathbb{R}^{D\times N\times (T\times 17)}$.
And $D,N,T,O$ are the embedding size, number of instances, number of frames, and number of pooled prompts, respectively. 

The pooling mechanism reduces the computational complexity and avoids out-of-memory (OOM) in the recognition decoder. 
The depth-wise mechanism, namely the mean operator $\mathcal{M}$ that reduces channels from $T\times20$ to $1$, is specifically designed for input flexibility across temporal and prompt dimensions. Consequently, a model trained on $T_1$ frames can perform inference on $T_2$ frames $(T_2\neq T_1)$ without requiring architectural modifications or weight changes. Similarly, the model is able to infer with either bounding box prompts alone or skeleton prompts alone, even when trained on both prompts.
Note that the instance dimension $N$ is not pooled and is required by relative identities encoding in $\S$\ref{sec:recognition_decoding}.

\subsection{Cross-Visual Decoding \label{sec:recognition_decoding}}
Leveraging the image embedding $\textbf{F}_I$, the prompt features $\textbf{F}_p$ and associated instance identities, PromptGAR takes the GAR class token $\textbf{X}_{gar}$ to decode the GAR logits $\textbf{Y}_{gar}$.

\begin{figure}[h]
\centering
\includegraphics[width=0.75\linewidth]{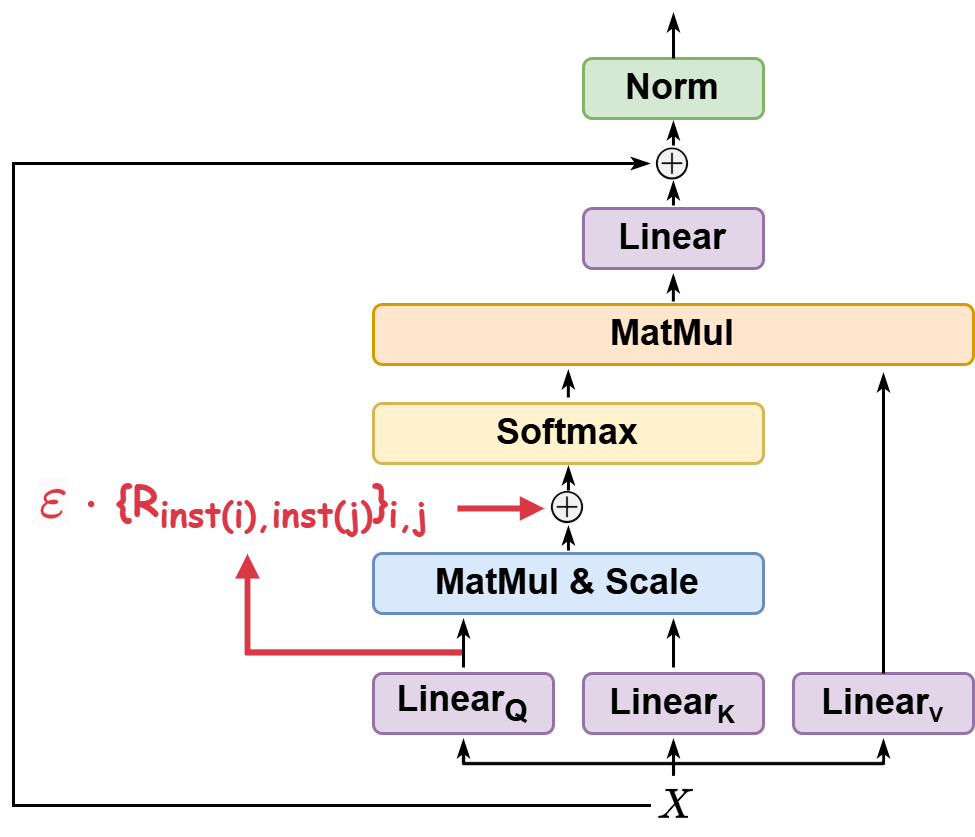}
\vspace{1.0em}
\caption{The \textbf{Relative Instance Attention} mechanism that incorporating a constant scale $\epsilon$ and relative instance identity embeddings $\mathcal{R}_{\text{inst}(i), \text{inst}(j)}$ in the attention block.}
\label{fig:rel_inst_self_attn}
\end{figure}

\textbf{Two-Way Transformer}.
For simplicity, we refer to these embeddings (not including
the image embedding) collectively as ``tokens''.
\begin{equation*}
    \begin{aligned}
        \texttt{images}&=\textbf{F}_I,\quad\texttt{tokens}=[\textbf{X}_{gar};\textbf{F}_{p}] \\
        \widehat{\textbf{F}}_I, [\widehat{\textbf{X}}_{gar};\widehat{\textbf{F}}_{p}] &= \text{Two-Way}(\texttt{images}, \texttt{tokens})
    \end{aligned}
\end{equation*}
Our two-way transformer is shown in Fig. \ref{fig:recognition_decoder}-(c), inspired by \cite{kirillov2023segment}, each layer performs 4 steps: 
(1) relative instance self-attention on tokens, 
(2) cross-attention from tokens to the image embedding,
(3) an MLP updates each token,
and (4) cross-attention from the image embedding to tokens.
The next layer takes the updated tokens and the updated image embedding from the previous layer. 
Input flexibility is well-achieved through the attention mechanism, which naturally handles inputs of different lengths. This allows our two-way transformer to process varying numbers of frames, types, or instances at inference time, regardless of those in training.

\textbf{Relative Instance Attention}. 
Existing research has demonstrated potential in modeling short-term temporal consistency via either optical flows \cite{pramono2020empowering, li2021groupformer, han2022dual} or joint motions \cite{zhou2022composer, li2025skeleton}. As these methods primarily focus on capturing motion between immediate time steps, the long-term consistency in the movement of objects or individuals is not well handled.
Also, the use of absolute instance \textit{ID} encoding to ensure consistency violates the principle of shift invariance \cite{lecun1989handwritten}. Namely, the interaction between two prompts becomes dependent on their arbitrary instance \textit{ID}s, even if they refer to the same underlying object or entity.
Inspired by relative positional embedding, we introduce the relative instance identity embedding to address this issue by focusing on whether two tokens belong to the same instance.

As illustrated in Fig. \ref{fig:rel_inst_self_attn}, we encode the relative instance information between the two input elements, $i$ and $j$, into embedding $\{R_{\text{inst}(i),\text{ inst}(j)}\}_{i,j}\in\mathbb{R}^D$, where $\text{inst}(i)$ and $\text{inst}(j)$ denote the instance \textit{ID} of element $i$ and $j$.
Notice that the GAR class token $\textbf{X}_{gar}$ is not included in $E^{(rel)}$.
$$\text{Attn}(Q,K,V)=\text{Softmax}\Big((QK^T)/\sqrt{d} + \epsilon\cdot E^{(rel)}\Big)V,$$
$$\text{where}\quad E^{(rel)}_{i,j}=Q_i\cdot \{R_{\text{inst}(i),\text{ inst}(j)}\}_{i,j}$$
Here, the number of token pairs belonging to different instances significantly outnumbers those belonging to the same instance. To address the imbalance problem, we assign a learnable embedding $R$ to pairs of tokens within the same instance, while setting the embedding to $0$ for pairs from different instances.
\begin{equation*}
    \{R_{\text{inst}(i),\text{ inst}(j)}\}_{i,j} = 
        \left\{ 
            \begin{aligned}
                0,\quad \text{inst}(i)&\ne\text{ inst}(j) \\
                R,\quad \text{inst}(i)&=\text{ inst}(j) \\
            \end{aligned}
        \right.
\end{equation*}
Due to the sparsity of $E^{(rel)}$, the scaling factor $\epsilon$ is a constant determined by experiments, with details in $\S$\ref{sec:ablation_studies}.
Note that relative instance embeddings cannot be applied to cross-attention between image embeddings and tokens, as image embeddings lack associated instance \textit{ID} information.

\textbf{GAR Head}.
In contrast to most approaches \cite{dosovitskiy2020image, kirillov2023segment, li2022mvitv2, yang2025x}, where \texttt{[class]} tokens are directly fed to MLP layers for logits prediction, we observe a unique challenge in GAR. While traditional methods align different prompts with distinct ground truth labels \cite{kirillov2023segment}, in GAR, various prompts correspond to the same ground truth label. 
As illustrated in Fig. \ref{fig:recognition_decoder}-(e), to encourage the network to utilize the information provided by the prompts, we add the class tokens with the averaged prompt features. Then a linear projection is employed to generate final logits for prediction.
$$\textbf{Y}_{gar}=\text{Linear}\Big(\widehat{\textbf{X}}_{gar} + \text{Mean}(\widehat{\textbf{F}}_p)\Big)$$
where $\text{Mean}(\widehat{\textbf{F}}_p)\in\mathbb{R}^D$ means taking average over $T\times O$. 
Our approach is also carefully designed to ensure input flexibility, even when the prompts present during the training time are absent during the inference time. During backpropagation, the gradients of $\widehat{\textbf{X}}_{gar}$ and $\text{Mean}(\widehat{\textbf{F}}_p)$ are identical, indicating they are optimized in the same direction and towards the same representation. Consequently, a well-trained model can function effectively using either $\widehat{\textbf{X}}_{gar}$ or $\text{Mean}(\widehat{\textbf{F}}_p)$ independently.

\section{Experiments}

\subsection{Experimental Setup}


\begin{table}[h!]
\small
\setlength\tabcolsep{3pt}

\makebox[\columnwidth][c]{
\centering
\begin{tabular}{l|ccccc|cc}
\hline
 & \multicolumn{5}{c|}{Prompt Types} & \multirow{2}{*}{\begin{tabular}[c]{@{}c@{}} Top1 \\ Acc\end{tabular}} & \multirow{2}{*}{\begin{tabular}[c]{@{}c@{}} Mean \\ Acc\end{tabular}} \\
\multirow{-2}{*}{Method} & RGB & Bbox & Kpt & Flow & Ball & & \\ \hline
ARG \cite{wu2019learning} & \ding{52} & \ding{52} &  &  &  & 90.7 & 91.0 \\
HiGCIN \cite{yan2020higcin} & \ding{52} & \ding{52} &  &  &  & 91.4 & 92.0 \\
DIN \cite{yuan2021spatio} & \ding{52} & \ding{52} &  &  &  & $92.7^\star$ & $92.8^\star$ \\ \hline
POGARS \cite{thilakarathne2022pose} &  &  & \ding{52} &  & \ding{52} & 93.9 & - \\
Composer \cite{zhou2022composer} &  &  & \ding{52} &  & \ding{52} & $93.6^\star$ & - \\
SkeleTR \cite{duan2023skeletr} &  &  & \ding{52} &  &  & 94.4 & - \\
MP-GCN \cite{li2025skeleton} &  &  & \ding{52} &  & \ding{52} & $95.0^\dagger$ & $95.0^\dagger$ \\ 
Bi-Causal \cite{zhang2024bi} &  &  & \ding{52} &  & \ding{52} & $95.8^\star$ & - \\  \hline
CRM \cite{azar2019convolutional} & \ding{52} & \ding{52} &  & \ding{52} &  & 93.0 & - \\
ActorFormer \cite{gavrilyuk2020actor} &  & \ding{52} & \ding{52} & \ding{52} &  & 94.4 & - \\
GIRN \cite{perez2022skeleton} & \ding{52} & \ding{52} & \ding{52} & \ding{52} &  & 94.0 & - \\
SACRF \cite{pramono2020empowering} & \ding{52} & \ding{52} & \ding{52} & \ding{52} &  & 95.0 & - \\
GroupFormer \cite{li2021groupformer} & \ding{52} & \ding{52} & \ding{52} & \ding{52} &  & 95.7 & - \\
Dual-AI \cite{han2022dual} & \ding{52} & \ding{52} &  & \ding{52} &  & 95.4 & - \\
KRGFormer \cite{pei2023key} & \ding{52} & \ding{52} & \ding{52} &  &  & 94.6 & 94.8 \\ \hline
\rowcolor[HTML]{dae8fc} 
\rule[-5pt]{0pt}{16pt}\textbf{PromptGAR} & \ding{52} & \ding{52} & \ding{52} &  &  & \textbf{96.0} & \textbf{96.3} \\ \hline
\end{tabular}
}

\caption{\textbf{Quantitative Comparisons in Volleyball Dataset}. 
($-$) denotes not reported in the paper,
($*$) reproduced from released codes, 
($\dagger$) reproduced with keypoint-ball results (late-fusion parts un-released), 
and unmarked other methods are not open-sourced.
}
\label{tab:volleyball_gar}
\end{table}


\begin{table}[h]
\small
\setlength\tabcolsep{3pt}

\makebox[\columnwidth][c]{
\centering
\begin{tabular}{l|ccccc|cc}
\hline
 & \multicolumn{5}{c|}{Prompt Types} & \multirow{2}{*}{\begin{tabular}[c]{@{}c@{}}Top1\\ Acc\end{tabular}}  & \multirow{2}{*}{\begin{tabular}[c]{@{}c@{}}Mean\\ Acc\end{tabular}} \\
\multirow{-2}{*}{Method} & RGB & Bbox & Kpt & Flow & Ball &  & \\ \hline
ARG \cite{wu2019learning} & \ding{52} & \ding{52} &  &  &  & 59.0 & 56.8 \\
ActorFormer \cite{gavrilyuk2020actor} & \ding{52} & \ding{52} &  & \ding{52} &  & 47.1 & 41.5 \\
SAM \cite{yan2020social} & \ding{52} & \ding{52} &  &  &  & 49.1 & 47.5 \\
DIN \cite{yuan2021spatio} & \ding{52} & \ding{52} &  &  &  & 61.6 & 56.0 \\
Dual-AI \cite{han2022dual} & \ding{52} & \ding{52} &  & \ding{52} &  & 58.1 & 50.2 \\
KRGFormer \cite{pei2023key} & \ding{52} & \ding{52} &  &  &  & 72.4 & 67.1 \\  \hline
MP-GCN \cite{li2025skeleton} &  &  & \ding{52} &  & \ding{52} & $75.8^\dagger$ & $72.0^\dagger$ \\ \hline
DFWSGAR \cite{kim2022detector} & \ding{52} &  &  &  &  & 75.8 & 71.2 \\ 
Flaming-Net \cite{nugroho2024flow} & \ding{52} &  &  & \ding{52} &  & 79.1 & 76.0 \\ \hline
\rowcolor[HTML]{dae8fc} 
\rule[-5pt]{0pt}{16pt}\textbf{PromptGAR}& \ding{52} & \ding{52} & \ding{52} &  &  & \textbf{80.6} & \textbf{76.9} \\ \hline
\end{tabular}
}

\caption{\textbf{Quantitative Comparisons in NBA Dataset}. ($\dagger$) reproduced with keypoint-ball results (late-fusion and multi-ensemble parts un-released), 
and unmarked methods are not open-sourced.}
\label{tab:nba_gar}
\end{table}

\textbf{Volleyball Dataset} \cite{ibrahim2016hierarchical} contains 3,493 clips for training and 1,337 clips for testing. Each clip has 41 frames and is labeled with one of eight group activities. Bounding box and skeleton annotations, provided separately by \cite{ibrahim2016hierarchical, zhou2022composer}, are available only for the central 16 frames of each clip. 
Consistent with prior works, we limit our analysis for central 16 frames to avoid potential interference from additional group activities present in the remaining half of clips. 


\textbf{NBA Dataset} \cite{yan2020social} includes 7,624 training clips and 1,548 testing clips. Each clip includes 72 frames and is categorized into nine group activity labels. 
Due to its increasing complexity, the NBA dataset demands special design and processing compared to the Volleyball dataset \cite{ibrahim2016hierarchical}.
Unlike Volleyball \cite{ibrahim2016hierarchical}, we use all 72 frames in NBA because key event frames are not centrally located. Also, NBA activities are much longer than Volleyball's. For example, `\textit{3p-fail-offensive}' requires recognizing shooter location, ball trajectory, and possession.

\textbf{Implementation details}. We utilize the MViTv2-Base \cite{li2022mvitv2} video encoder with a $224\times224$ input resolution. For the Volleyball dataset \cite{ibrahim2016hierarchical}, training and testing use the center 16 frames. The NBA dataset \cite{yan2020social} is trained on 56 uniformly sampled frames and tested on all 72. The recognition decoder consists of an eight-layer stack of two-way transformers. All models are trained on 4 A100-80GB GPUs, where the Volleyball dataset use a batch size of 64, and the NBA dataset employ 24. Training runs for 200 epochs, using an AdamW optimizer and a Cosine Annealing learning rate scheduler. The initial learning rates are $2\times10^{-4}$ for Volleyball and $7.5\times10^{-5}$ for NBA.

\subsection{Group Activity Recognition}

\textbf{Volleyball Dataset}.
In Tab. \ref{tab:volleyball_gar}, we compare our method to the group activity recognition methods that rely on: 
(a) RGB frames, like DIN \cite{yuan2021spatio}, which suffers from limited performance; 
(b) skeletons and ball trajectories, like Composer \cite{zhou2022composer} and MP-GCN \cite{li2025skeleton}, which requires ball trajectories as extra inputs; 
(c) combined visual prompts, like GroupFormer \cite{li2021groupformer} and Dual-AI \cite{han2022dual}, which takes computationally expensive optical flows as inputs and thus sensitive to input frame rates.
In contrast, our PromptGAR, without the drawbacks of those other visual prompt inputs, still produces 96.0\% top1 accuracy and 96.3\% mean accuracy, achieving considerable improvements across different baseline models. Also, ACCG \cite{xie2023actor} can not be evaluated due to unreleased codes.

\textbf{NBA Dataset}.
Tab. \ref{tab:nba_gar} shows that PromptGAR gets competitive performance over previous methods on the challenging NBA dataset \cite{yan2020social}. 
Prior approaches often face limitations.
Older models like SAM \cite{yan2020social} and Dual-AI \cite{han2022dual} accept only raw videos and bounding boxes, which restricts their performance;
Other methods, such as MP-GCN \cite{li2025skeleton}, require specialized inputs like ball trajectories, thereby limiting their applicability to broader, non-sports GAR tasks. 
Additionally, weakly-supervised techniques like DFWSGAR \cite{kim2022detector} and Flaming-Net \cite{nugroho2024flow} introduce considerable training complexity.
In contrast, our PromptGAR, without these drawbacks, still produces 80.6\% top1 accuracy and 76.9\% mean accuracy, reflecting substantial advancements over various baseline models.

\subsection{Input Flexibility}

\textbf{Flexible Prompts}.
Tab. \ref{tab:less_prompts} demonstrates PromptGAR's resilience to reduced prompt information. Using only skeleton data, it loses a negligible 0.3\% in accuracy. When relying solely on bounding boxes, it achieves the same accuracy as GroupFormer \cite{li2021groupformer}, but without retraining. Even with only RGB input, the model's accuracy remains reasonable, only 2.1\% lower than with full prompts. This shows PromptGAR's robustness to maintain high performance even with significantly reduced prompt inputs without retraining.

\begin{table}[]
\small
\setlength\tabcolsep{2pt}

\makebox[\columnwidth][c]{
\centering
\begin{tabular}{l|cccc|c|cc}
\hline
\rule[-3pt]{0pt}{12pt} & \multicolumn{4}{c|}{Prompt Types} &  &  &  \\
\multirow{-2}{*}{Method} & RGB & Bbox & Kpt & Flow & \multirow{-2}{*}{Re-train} & \multirow{-2}{*}{\begin{tabular}[c]{@{}c@{}}Top1\\ Acc\end{tabular}} & \multirow{-2}{*}{\begin{tabular}[c]{@{}c@{}}Mean\\ Acc\end{tabular}} \\ \hline
 & \ding{51} & \ding{51} & \ding{51} & \ding{51} &  & 95.7 & - \\
 & \ding{51} & \ding{51} &  & \ding{51} &  & 94.9 & - \\
\multirow{-3}{*}{GroupFormer \cite{li2021groupformer}} & \ding{51} & \ding{51} &  &  & \multirow{-3}{*}{\ding{51}} & 94.1 & - \\ \hline
\rule[-4pt]{0pt}{13pt} & \cellcolor[HTML]{e6e6e6}\ding{51} & \cellcolor[HTML]{e6e6e6}\ding{51} & \cellcolor[HTML]{e6e6e6}\ding{51} & \cellcolor[HTML]{e6e6e6} &  & \cellcolor[HTML]{e6e6e6}\textbf{96.0} & \cellcolor[HTML]{e6e6e6}\textbf{96.3} \\
 & \ding{51} &  & \ding{51} &  &  & 95.7 & 95.8 \\
 & \ding{51} & \ding{51} &  &  &  & 94.1 & 94.3 \\
\multirow{-4}{*}{\textbf{PromptGAR}} & \ding{51} &  &  &  & \multirow{-4}{*}{\ding{53}} & 93.9 & 94.4 \\ \hline
\end{tabular}
}

\caption{\textbf{Flexible Prompts}.
In the Volleyball dataset, we achieve remarkable performance under diverse prompts without retraining. 
}
\label{tab:less_prompts}
\end{table}

\begin{table}[]
\small
\setlength\tabcolsep{2.5pt}

\makebox[\columnwidth][c]{
\centering
\begin{tabular}{l|cc|cc|c|cc}
\hline
 & \multicolumn{2}{c|}{Train} & \multicolumn{2}{c|}{Test} &  &  \\
\multirow{-2}{*}{Method} & $T$ & Sampling & $T$ & Sampling & \multirow{-2}{*}{Re-train} & \multirow{-2}{*}{\begin{tabular}[c]{@{}c@{}}Top1\\ Acc\end{tabular}} & \multirow{-2}{*}{\begin{tabular}[c]{@{}c@{}}Mean\\ Acc\end{tabular}} \\ \hline
 & 18 & stride 1 & 18 & stride 1 &  & 73.3 & 68.4 \\
\multirow{-2}{*}{MP-GCN \cite{li2025skeleton}} & 72 & stride 1 & 72 & stride 1 & \multirow{-2}{*}{\ding{51}} & 75.8 & 72.0 \\ \hline
 &  &  & 36 & stride 2 &  & 75.4 & 71.9 \\
 &  &  & 56 & stride 1 &  & 75.0 & 71.0 \\
 &  &  & 64 & stride 1 &  & 78.4 & 74.1 \\
\rule[-4pt]{0pt}{13pt}\multirow{-4}{*}{\textbf{PromptGAR}} & \multirow{-4}{*}{56} & \multirow{-4}{*}{uniform} & \cellcolor[HTML]{e6e6e6}72 & \cellcolor[HTML]{e6e6e6}stride 1 & \multirow{-4}{*}{\ding{53}} & \cellcolor[HTML]{e6e6e6} \textbf{80.6} & \cellcolor[HTML]{e6e6e6}\textbf{76.9} \\ \hline
\end{tabular}
}

\caption{\textbf{Flexible Frames}. For the NBA dataset, frames are all sampled at the center of each video clip. PromptGAR's validation process also uses $T=72$ to select the optimal checkpoint.}
\label{tab:flexible_frames}
\end{table}
\begin{table}[]
\small
\setlength\tabcolsep{4pt}
\centering

\begin{tabular}{l|c|c|cc}
\hline
\rule[-5pt]{0pt}{16pt}Method & \# Instances & Re-train & Top1 Acc & Mean Acc \\ \hline
\rule[-5pt]{0pt}{16pt} & \cellcolor[HTML]{e6e6e6}12 &  & \cellcolor[HTML]{e6e6e6}\textbf{96.0} & \cellcolor[HTML]{e6e6e6}\textbf{96.3} \\
 & 10 &  & 95.4 & 95.6 \\
 & 5 &  & 94.5 & 94.8 \\
\multirow{-4}{*}{\textbf{PromptGAR}} & 3 & \multirow{-4}{*}{\ding{53}} & 94.3 & 94.6 \\ \hline
\end{tabular}

\caption{\textbf{Flexible Instances}. In the Volleyball dataset, instances are randomly deleted, and results are averaged over three trials.}
\label{tab:fewer_instances}
\end{table}

\textbf{Flexible Frames}.
Tab. \ref{tab:flexible_frames} exhibits PromptGAR's adaptability to various frame sampling configurations. Although trained on 56 frames with the NBA dataset, it can effectively handle both shorter and longer sequences at test time. Specifically, it achieves 75.4\% accuracy when tested with 36 frames (stride 2), nearly matching MP-GCN's performance, which needs to be trained and tested on the full 72 frames specifically. Moreoever, PromptGAR  also outperforms MP-GCN when tested on all 72 frames, even though MP-GCN is optimized for this exact frame count.
Furthermore, the effective rollout length during testing is crucial. When testing with 56 frames and a stride of 1, we see a 0.4\% accuracy decrease compared to 36 frames with a stride of 2, due to the shorter rollout length (56 vs. 72).
Those results demonstrate PromptGAR’s robust ability to handle diverse frame inputs without the need for retraining, a clear advantage over methods requiring retraining.

\textbf{Flexible Instances}.
Tab. \ref{tab:fewer_instances} showcases that PromptGAR can maintain high performance even when tested with fewer instances than it was trained on, without requiring retraining. To ensure a fair comparison, instances were randomly selected, and results were averaged over three experiments. Reducing from 12 instances to 10 results in only a 0.6\% accuracy drop, and even when reduced to just 3 instances, the drop is still manageable at 1.7\%. This highlights PromptGAR’s ability to function with reduced input, without the needs for retraining.

\subsection{Actor Consistency \label{sec:actor_consistency}}


\begin{table}[b]
\small

\makebox[\columnwidth][c]{
\centering
\begin{tabular}{l|c|c|cc}
\hline
Method & \begin{tabular}[c]{@{}c@{}}Shuffle Input \\ Actor Order\end{tabular} & Re-train & \begin{tabular}[c]{@{}c@{}}Top1 \\ Acc\end{tabular} & \begin{tabular}[c]{@{}c@{}}Mean \\ Acc\end{tabular} \\ \hline
\multirow{2}{*}{MP-GCN \cite{li2025skeleton}} & \ding{53} & \multirow{2}{*}{\ding{53}} & 75.8 & 72.0 \\
 & \ding{51} &  & 74.7 & 71.2 \\ \hline
\multirow{2}{*}{\textbf{PromptGAR}} & \ding{53} & \multirow{2}{*}{\ding{53}} & 80.6 & 76.9 \\
 & \ding{51} &  & \cellcolor[HTML]{EFEFEF}\textbf{80.6} & \cellcolor[HTML]{EFEFEF}\textbf{76.9} \\ \hline
\end{tabular}
}
\caption{\textbf{Actor Order Invariance}. For NBA dataset, input actor order is randomly shuffled, and results are averaged over five trials.}
\label{tab:actor_consist}
\end{table}

\begin{table}[b]
\small
\setlength\tabcolsep{3pt}

\makebox[\columnwidth][c]{
\centering
\begin{tabular}{l|c|c|cc}
\hline
Method & \begin{tabular}[c]{@{}c@{}}Relative Instance \\ Identity Embeddings\end{tabular} & Re-train & \begin{tabular}[c]{@{}c@{}}Top1\\ Acc\end{tabular} & \begin{tabular}[c]{@{}c@{}}Mean\\ Acc\end{tabular} \\ \hline
\rule[-5pt]{0pt}{12pt}\multirow{2.2}{*}{\textbf{PromptGAR}} & \ding{51} & & \cellcolor[HTML]{EFEFEF}\textbf{96.0} & \cellcolor[HTML]{EFEFEF}\textbf{96.3} \\ 
  & \ding{53} & \multirow{-2}{*}{\ding{53}} &93.8 & 94.1 \\ \hline
\end{tabular}
}
\caption{\textbf{Necessity of Instance Identities}. In Volleyball dataset, relative instance identity embeddings are included or omitted, using prompts from bounding boxes, skeletons, and RGB videos.}
\label{tab:flex_prompt_wo_instids}
\end{table}

\textbf{Actor Order Invariance}. 
Prompt inputs for the testing dataset are formatted as a tensor of shape $(N, T, M, J, C)$, representing the number of videos, frames, actors, point types, and coordinates, respectively. 
Bounding boxes and skeletal keypoints belonging to the same instance \textit{ID} share the same index along the $M$ (actor) axis.
The actor order along this $M$ axis is usually fixed in conventional evaluations, but we introduce a rigorous test: we randomly shuffle the $M$ axis and evaluate the same model. 
This procedure is repeated five times, and the averaged performance is presented in Tab. \ref{tab:actor_consist}. 
Our results demonstrate that after shuffling, MP-GCN \cite{li2025skeleton} shows a slight performance degradation, but our model maintains identical performance numbers. 
This is because current GAR methods \cite{zhou2022composer, li2025skeleton, gavrilyuk2020actor, li2021groupformer, han2022dual} fix that order for both validation and testing, choosing the best checkpoint based on performance with that specific player order.
In contrast, our model incorporates a novel relative instance attention mechanism that encodes only whether two actors are the same or not. 
This design ensures our model remains unaffected by actor order in real-world scenarios, thereby enhancing its robustness.

\textbf{Necessity of Instance Identities}. 
As shown in Tab. \ref{tab:flex_prompt_wo_instids}, there is a performance drop without instance identities. 
This is because, without instance identities, the model struggles to maintain actor consistency and reliably associate individual features across different frames. 
This also proves the necessity of our instance identities.


\begin{table}[]
\small
\setlength\tabcolsep{4pt}
\centering

\begin{tabular}{lccc}
\hline
\rule[-5pt]{0pt}{16pt}Method & Re-train & Top1 Acc & Mean Acc \\ \hline
\rule[-3pt]{0pt}{12pt}\cellcolor[HTML]{e6e6e6}\textbf{PromptGAR} & \cellcolor[HTML]{e6e6e6}\ding{51} & \cellcolor[HTML]{e6e6e6}\textbf{95.8} & \cellcolor[HTML]{e6e6e6}\textbf{96.0} \\
\textit{wo inst IDs} & \ding{51} & 95.4 & 95.6 \\
\textit{wo prompt tokens in head} & \ding{51} & 95.1 & 95.4 \\
\textit{no prompts} & \ding{51} & 94.3 & 94.7 \\ \hline
\end{tabular}
\vspace{1.0em}
\caption{\textbf{Effectiveness of Novel Modules}.
Experiments are conducted using smaller models with $O=16$ in Volleyball.
Ablation studies included: 
`\textit{wo inst IDs}', where relative instance attention is replaced with regular self-attention; 
`\textit{wo prompt tokens in head}', where only the GAR class token $\textbf{X}_{gar}$ is used for the GAR head; 
and `\textit{no prompts}', where the recognition decoder receives only the GAR class token $\textbf{X}_{gar}$ and RGB features $\textbf{F}_I$ as inputs.
}
\label{tab:ablation_studies}
\end{table}

\begingroup
\captionsetup{font=footnotesize}
\begin{table}
\centering
\begin{minipage}[t]{0.5\columnwidth}
\centering
\footnotesize
\setlength\tabcolsep{3pt}
\begin{tabular}{c|c|cc}
\hline
$O$ & Memory & Top1 Acc & Mean Acc \\ \hline
16 & 40 \textit{GB} & 95.8 & 96.0 \\
\rowcolor[HTML]{e6e6e6} 
\rule[-3pt]{0pt}{12pt}48 & 72 \textit{GB} & \textbf{96.0} & \textbf{96.3} \\
56 & \textit{OOM} & - & - \\ \hline
\end{tabular}
\vspace{0.08in}
{\footnotesize
\caption{\textbf{\# of Pooled Prompts $O$}.}
\label{tab:pooling_prompts}
}
\end{minipage}%
\hfill
\begin{minipage}[t]{0.5\columnwidth}
\centering
\footnotesize
\setlength\tabcolsep{3pt}
\begin{tabular}{c|cc}
\hline
$\epsilon$ & Top1 Acc & Mean Acc \\ \hline
0 & 95.4 & 95.6 \\
5 & 95.6 & 95.9 \\
\rowcolor[HTML]{e6e6e6} 
\rule[-3pt]{0pt}{12pt}10 &  \textbf{95.8} & \textbf{96.0} \\ \hline
\end{tabular}
\vspace{0.09in}
{\footnotesize
\caption{\textbf{Relative Instance Scale $\epsilon$}.}
\label{tab:rel_inst_scale}
}
\end{minipage}
\end{table}
\endgroup

\subsection{Ablation Studies \label{sec:ablation_studies}}

Tab. \ref{tab:ablation_studies} details the impact of our novel modules on performance, starting with a baseline of 96.0\% top-1 accuracy on the Volleyball dataset. 
(a) \textbf{Relative instance attention}. Eliminating instance \textit{ID}s during training and testing resulted in a 0.6\% drop in top-1 accuracy. This confirms that instance \textit{ID}s are vital for encoding long-term consistency, enhancing performance. Furthermore, relative instance attention also accelerated training convergence.
(b) \textbf{Prompt tokens in head}. Removing prompt tokens $\widehat{\textbf{F}}_p$ from the head, leaving only the GAR class token $\widehat{\textbf{X}}_{gar}$, led to a 0.3\% accuracy decrease. This highlights the necessity of explicitly incorporating prompt tokens. Without them, the model struggles to integrate diverse visual prompt features, as the GAR class token alone sufficiently captures RGB information.
(c) \textbf{No prompts}. Disabling the entire prompt encoder, using only the GAR class token $\widehat{\textbf{X}}_{gar}$ as input for the recognition decoder and the head, further reduced accuracy by 0.8\%. This underscores the contribution of visual prompt inputs in PromptGAR.
Crucially, even without any prompts during training and testing, our model still compares favorably with other prompt-based baseline methods. Specifically, it achieves a 1.6\% higher accuracy than DIN, which utilizes both RGB frames and bounding boxes, and a 0.7\% higher accuracy than Composer, which uses skeletons and ball trajectories as additional information. This demonstrates the superior ability of our MViTv2 architecture to process spatial-temporal information within only the RGB features compared to other backbones that process the video frame-by-frame separately.

\textbf{Number of Pooled Prompts}.
Tab. \ref{tab:pooling_prompts} explores the impact of the number of pooled prompt channels ($O$) on performance. The prompt encoder's output has the shape $D\times N\times O$, where $D, N, O$ are the feature dimension, the number of instances, and the pooled prompt channels after pooling along temporal and prompt type dimensions, repectively.
A smaller $O$ value reduces the representativeness and expressiveness of each actor's key information. As shown in the table, increasing $O$ from 16 to 48 improves top1 accuracy by 0.2\%. However, this increase quadratically raises the computational cost and CUDA memory usage of the subsequent recognition decoder module.
Specifically, the CUDA memory usage for $O=48$ is 32 GB higher than for $O=16$. Further increasing $O$ to 56 would exceed the 80 GB limit of an A100 GPU's CUDA memory, resulting in an out-of-memory error. Therefore, while a higher $O$ value improves performance, a balance must be struck to avoid excessive computational demands and memory limitations.

\textbf{Relative Instance Scale}.
Tab. \ref{tab:rel_inst_scale} details the process of determining the optimal relative instance scale, denoted as $\epsilon$. This value serves as a coefficient for the relative instance weights, which are added to the standard attention weights. Due to the inherent sparsity of the relative instance weight matrix $E^{(rel)}$, a small $\epsilon$ value would diminish the impact of instance \textit{ID}s during training. Conversely, an excessively large $\epsilon$ would overpower the standard attention weights, hindering the GAR class token's ability to learn effectively from the prompt tokens.
The table reveals that setting $\epsilon$ to 10 yields a 0.4\% improvement in top-1 accuracy compared to setting it to 0 (effectively disabling relative instance attention) and a 0.2\% improvement compared to setting it to 5. Therefore, we selected 10 as the final $\epsilon$ value.
Notice that these experiments are conducted with a pooled prompt channel count ($O$) of 16.

\section{Limitations}

Despite its competitive GAR accuracy and input flexibility, PromptGAR faces limitations with input resolution.
The input resolution is restricted to $224\times224$, which is significantly lower than $1080\times720$ in other methods \cite{yuan2021spatio, li2021groupformer, han2022dual}.


\section{Conclusion}
In this work, we present PromptGAR, a novel framework that addresses the limitations of current Group Activity Recognition (GAR) approaches by leveraging diverse visual prompts to achieve both input flexibility and high recognition accuracy. Extensive experiments show that PromptGAR allows the input flexibility across prompts, frames, and instances without the need for retraining. Furthermore, we introduce a novel relative instance attention mechanism that directly encodes instance identities, ensuring actor consistency. Extensive evaluation validates the effectiveness and applicability of our model.


{
    \small
    \bibliographystyle{ieeenat_fullname}
    \bibliography{main}
}

\clearpage
\newpage
\maketitlesupplementary
\setcounter{page}{1}
\setcounter{section}{0}


\begin{table}[h]
\footnotesize
\setlength\tabcolsep{2pt}

\makebox[\columnwidth][c]{
\centering
\begin{tabular}{l|cccc|cc|c|cc}
\hline
\rule[-3pt]{0pt}{12pt}\multirow{2}{*}{Method} & \multicolumn{4}{c|}{Prompt Types} & \multicolumn{2}{c|}{\cellcolor[HTML]{e6e6e6}\textbf{Head Inputs}} & \multirow{2}{*}{Re-train} & \multirow{2}{*}{Top1} & \multirow{2}{*}{Mean} \\
 & RGB & Bbox & Kpt & InstID & \cellcolor[HTML]{e6e6e6}\boldmath$\widehat{X}_{gar}$ & \cellcolor[HTML]{e6e6e6}\boldmath$\widehat{F}_p$ &  &  &  \\ \hline
\rule[-3pt]{0pt}{12pt}\multirow{10}{*}{\rule[-5pt]{0pt}{16pt}\textbf{PromptGAR}} & \cellcolor[HTML]{e6e6e6}\ding{51} & \cellcolor[HTML]{e6e6e6}\ding{51} & \cellcolor[HTML]{e6e6e6}\ding{51} & \cellcolor[HTML]{e6e6e6}\ding{51} & \cellcolor[HTML]{e6e6e6}\ding{51} & \cellcolor[HTML]{e6e6e6}\ding{51} & \multirow{10}{*}{\ding{53}} & \cellcolor[HTML]{e6e6e6}\textbf{96.0} & \cellcolor[HTML]{e6e6e6}\textbf{96.3} \\
 & \ding{51} &  & \ding{51} & \ding{51} & \cellcolor[HTML]{e6e6e6}\ding{51} & \cellcolor[HTML]{e6e6e6}\ding{51} &  & 95.7 & 95.8 \\
 & \ding{51} & \ding{51} &  & \ding{51} & \cellcolor[HTML]{e6e6e6}\ding{51} & \cellcolor[HTML]{e6e6e6}\ding{51} &  & 94.1 & 94.3 \\ \cline{2-5} \cline{9-10} 
 & \ding{51} & \ding{51} & \ding{51} & \ding{51} & \cellcolor[HTML]{e6e6e6}\ding{51} & \cellcolor[HTML]{e6e6e6} &  & 95.8 & 96.0 \\
 & \ding{51} &  & \ding{51} & \ding{51} & \cellcolor[HTML]{e6e6e6}\ding{51} & \cellcolor[HTML]{e6e6e6} &  & 95.5 & 95.7 \\
 & \ding{51} & \ding{51} &  & \ding{51} & \cellcolor[HTML]{e6e6e6}\ding{51} & \cellcolor[HTML]{e6e6e6} &  & 94.0 & 94.2 \\
 & \ding{51} &  &  &   & \cellcolor[HTML]{e6e6e6}\ding{51} & \cellcolor[HTML]{e6e6e6} &  & 93.9 & 94.4 \\ \cline{2-5} \cline{9-10} 
 & \ding{51} & \ding{51} & \ding{51} & \ding{51} & \cellcolor[HTML]{e6e6e6} & \cellcolor[HTML]{e6e6e6}\ding{51} &  & 95.2 & 95.6 \\
 & \ding{51} &  & \ding{51} & \ding{51} & \cellcolor[HTML]{e6e6e6} & \cellcolor[HTML]{e6e6e6}\ding{51} &  & 95.2 & 95.6 \\
 & \ding{51} & \ding{51} &  & \ding{51} & \cellcolor[HTML]{e6e6e6} & \cellcolor[HTML]{e6e6e6}\ding{51} &  & 93.4 & 94.0 \\ \hline
\end{tabular}
}

\caption{\textbf{Ablation on Head Input Tokens}.}
\label{tab:head_inputs}
\end{table}


\begin{table}[h]
\footnotesize
\setlength\tabcolsep{2pt}

\makebox[\columnwidth][c]{
\centering
\begin{tabular}{l|cccc|cccc|cc}
\hline
  & \multicolumn{4}{c|}{Train}& \multicolumn{4}{c|}{Test}   &  &  \\
\multirow{-2}{*}{Method}    & RGB & Bbox& Kpt & \multicolumn{1}{l|}{InstID} & RGB    & Bbox   & Kpt    & \multicolumn{1}{l|}{InstID}  & \multirow{-2}{*}{\begin{tabular}[c]{@{}c@{}}Top1\\ Acc\end{tabular}} & \multirow{-2}{*}{\begin{tabular}[c]{@{}c@{}}Mean\\ Acc\end{tabular}} \\ \hline
  &&&&   & \ding{51} & \ding{51} & \ding{51} & \ding{51}  & \cellcolor[HTML]{EFEFEF}\textbf{96.0} & \cellcolor[HTML]{EFEFEF}\textbf{96.3} \\
  & \multirow{-2}{*}{\ding{51}} & \multirow{-2}{*}{\ding{51}} & \multirow{-2}{*}{\ding{51}} & \multirow{-2}{*}{\ding{51}} & \ding{51} &   & \ding{51} & \ding{51}  & 95.7   & 95.8   \\ \cline{2-11}
\rule[-5pt]{0pt}{16pt} & \ding{51}    & \ding{51}    && \ding{51} & \ding{51} & \ding{51} &   & \ding{51} & 95.4   & 95.8  \\ \cline{2-11}
\rule[-5pt]{0pt}{16pt}\multirow{-5}{*}{\textbf{PromptGAR}} & \ding{51}    &    && & \ding{51} & &   &  & 94.3   & 94.7  \\ \hline
\end{tabular}
}
\caption{\textbf{Low Reliance on Skeletons}. }
\label{tab:low_kpt_reliance}
\end{table}

\section{More on Implementation Details}

\textbf{Data Augmentations}. Following MViTv2 in action recognition \cite{li2022mvitv2}, we apply a comprehensive set of augmentations for group activity recognition (GAR): Random Augment \cite{cubuk2020randaugment}, Random Resized Crop, Flip, Random Erasing \cite{zhong2020random}, CutMix \cite{yun2019cutmix}, and MixUp \cite{zhang2017mixup}. 

\textbf{Volleyball Dataset} \cite{ibrahim2016hierarchical}. Since Volleyball only contains annotations with central 16 frames, we replace the depth-wise prompt pooling with a single MLP layer projecting channels from $T\times20$ to the number of pooled prompts $O$.

\textbf{NBA Dataset} \cite{yan2020social}. We regenerated annotations due to inaccuracies in SAM \cite{yan2020social} and MP-GCN \cite{li2025skeleton}, such as background audience false positives, player ID switching, and ID reassignment upon reappearance. We used MOTIP \cite{gao2024multiple} for robust player tracking, optimized for sports data, and RTMPose \cite{jiang2023rtmpose} for accurate skeleton generation, handling occlusions effectively.

\section{Additional Ablation Studies}


\begin{figure*}[h]
\centering
\begin{minipage}[t]{1\columnwidth}
\includegraphics[width=1\linewidth]{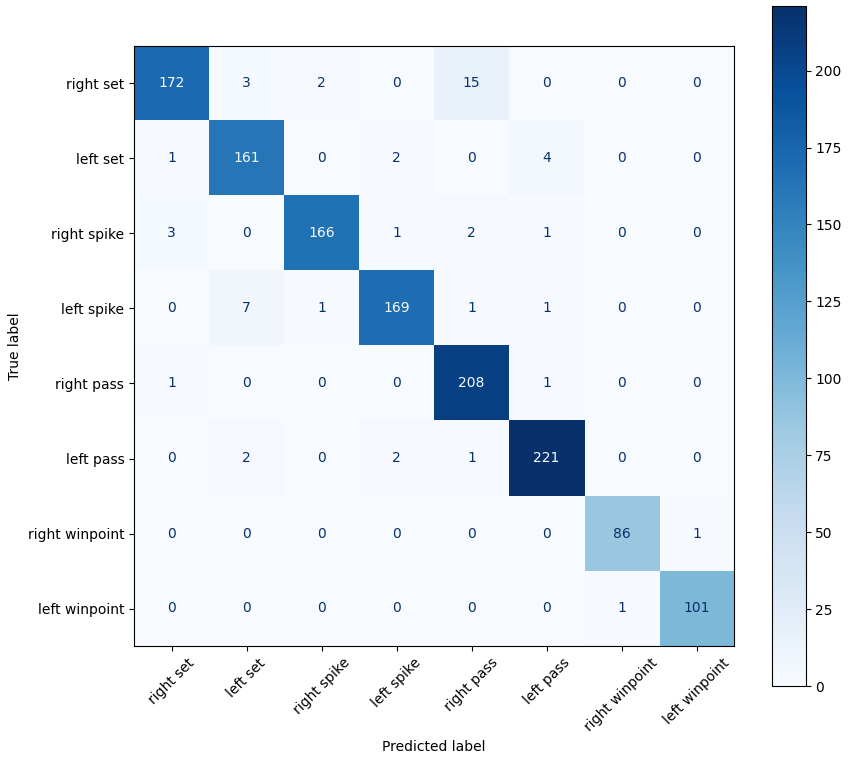}
\caption{\textbf{Volleyball Confusion Matrix}}
\label{fig:volleyball_cm}
\end{minipage}%
\hfill
\begin{minipage}[t]{1\columnwidth}
\centering
\includegraphics[width=1\linewidth]{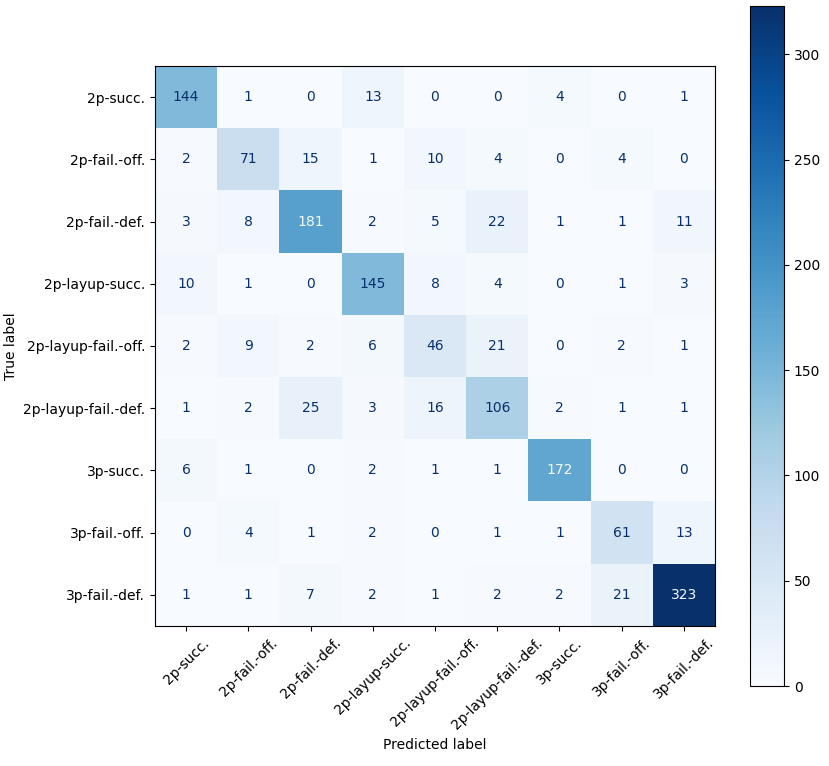}
\caption{\textbf{NBA Confusion Matrix}}
\label{fig:nba_cm}
\end{minipage}%
\end{figure*}


\textbf{Head Inputs}.
Tab. \ref{tab:head_inputs} indicates demonstrates the significance of prompt token inputs within the head, even when dealing with diverse visual prompts. We compared performance using a model trained on full prompts, varying the head inputs during testing.
Under full prompt inputs, using only the GAR class token $\widehat{\textbf{X}}_{gar}$ as the head input resulted in a minor 0.2\% performance drop, while using only the prompt tokens $\widehat{\textbf{F}}_p$ led to a 0.8\% drop. This shows that both inputs are crucial for PromptGAR's high performance.
Even with reduced prompt inputs, both $\widehat{\textbf{X}}_{gar}$ only and $\widehat{\textbf{F}}_p$ only still yielded reasonable results. 
Notably, when using only prompt tokens $\widehat{\textbf{F}}_p$, the performance of full prompt inputs was similar to the one with skeleton-only inputs. This suggests that when RGB features don't directly influence the final prediction, bounding box information is effectively captured within the skeleton data.
In summary, this table showcases the flexibility of head inputs across various visual prompts and underscores the importance of both the RGB feature representation $\widehat{\textbf{X}}_{gar}$ and the prompt feature representation $\widehat{\textbf{F}}_p$ for optimal performance.

\textbf{Low Reliance on Skeletons}.
Tab. \ref{tab:low_kpt_reliance}, our model has competitive performance even when trained without skeletal data. 
This outcome indicates that our model does not heavily rely on skeletal information.

\section{Analysis}

\textbf{Quantitative Results}.
Fig. \ref{fig:volleyball_cm} - \ref{fig:nba_cm} shows the confusion matrix of Volleyball and NBA, respectively. 
(a) For Volleyball, while PromptGAR demonstrates high accuracy on the Volleyball dataset, a common misclassification occurs where `\textit{right-set}' is mistaken for `\textit{right-pass}'. This error is attributed to the inherent similarity in player actions and positions, echoing findings in previous research \cite{zhou2022composer}.
(b) For NBA, errors highlight the challenge of distinguishing between highly similar actions. Specifically, differentiating `\textit{offensive}' and `\textit{defensive}' requires nuanced visual analysis of player uniform differences and rebound outcomes. Similarly, distinguishing `\textit{2p-fail}' from `\textit{2p-layup-fail}' demands attention to shooting position and posture near the rim.

\end{document}